\documentclass{svjour3}  

\smartqed  
\usepackage{mathptmx}      
\usepackage{amsfonts,amssymb,amsmath,amsgen,amsopn,amsbsy,theorem,graphicx,epsfig}

\usepackage{latexsym}
\usepackage{multirow}
\usepackage[round]{natbib}
\usepackage{array}

\begin{document}
\title{Modeling of a Quadruped Robot with Spine Joints and Full-Dynamics Simulation Environment Construction
}


\author{Omer Kemal Adak        \and
        Beste Bahceci      \and
        Kemalettin Erbatur
}


\institute{Omer Kemal Adak \at
              Faculty of Engineering and Natural Sciences, Sabanci University \\
              \email{omerkemal@sabanciuniv.com}           
           \and
          Beste Bahceci \at
              Faculty of Engineering and Natural Sciences, Sabanci University \\
              \email{bestebahceci@sabanciuniv.com}
         \and
          Kemalettin Erbatur \at
              Faculty of Engineering and Natural Sciences, Sabanci University \\
              \email{erbatur@sabanciuniv.com} 
}
\date{Received: date / Accepted: date}

\maketitle
\begin{abstract}
     This paper presents modeling and simulation of a spined quadruped robot. Extended literature survey is employed and spine joints researches of the quadruped robots are classified. Most of the researchers execute simplified quadruped robot models in their simulations. This survey reveals the need for the full-body spined quadruped simulation environment. First, the kinematics and dynamics modeling of the active spined quadruped robot is obtained. Since quadruped robots are floating-base robots, all derivations are performed with respect to an inertial frame. The motion equations are acquired by the Lagrangian approach. The simulation environment is constructed in the MATLAB/Simulink platform, considering its rich library,  powerful solvers, and suitable and resilient environment in integrating controllers. The computation speed of the simulation environment is increased by using optimized MATLAB functions. Precise and accurate contact model is utilized in the simulation environment. We foreseen that the provided full-dynamics simulation environment will be helpful for further spine joint studies on the quadruped robot field.
\end{abstract}

\section{Introduction}
\label{intro}
The legged robots have improved maneuverability and terrain adaptability owing to their articulated structure. They can work quickly and efficiently in rugged, fragmented, and unstable terrains that are inaccessible to wheeled or tracked vehicles \citep{raibert1986}. They have a lot of potential when it comes to employing specific tasks in diverse environments. Mine-sweeping, discovery, logistic transport, search and rescue operations, disaster zones, space, and military applications are all expected to use legged robotics in the future \citep{mahapatra}. The quadruped robot provides a valuable compromise between maneuverability and complexity of configuration and control in the field of legged robotics \citep{zhuang}. As a result, quadruped robots are a popular topic among robotic researchers.

In the last few decades, a vast number of quadruped robots have been manufactured to investigate the stable locomotion of legged systems in outdoor environment applications. The majority of them have shown remarkable performance with rigid bodies \citep{scout,tekken,bigdog,hyq,hyq2max,anymal}. These quadruped robots successfully carried out different gaits, passed across rugged terrains and executed challenging tasks. However, the motion of rigid-body quadruped robots is not sufficiently smooth compared to quadruped animals. 

When you look at running hoofed animals, there is a unique structure linking their front and back legs which is a flexible spine. Quadruped animals have evolved to run in several ways in nature, and the spine's work has specialized as a result. For instance, the duty of the spine is related to shock absorption, acceleration, and velocity in dynamic motion animals \citep{gambaryan}. However, these specializations are not sufficient to completely define the general functions of the spine.  

According to biomechanical studies, a flexible spine increases leg mobility. The spine acts as a kinematic extension of the leg in certain areas. Leg length is effectively increased as a result of this anatomical adaptation. Gray addressed how stride length increased by spinal expansion in quadruped animals \citep{gray2013,english}. Speed is thought to be increased by increasing stride length.

The components that create the spine and legs are serially linked kinematically. Motions at the foot derive from the total number of the motions of all the joints in the link, from the spine to the foot. An elevated leg velocity benefits from serially conveying motion on the foot by both spine and leg muscles, enhancing the speed of the animal \citep{hildebrand1959}. The flexion and expansion of the animal body contribute to increasing the animal's pace by allowing more distance to be covered during the aerial period per stride \citep{hildebrand1959}. Additionally, the massive flexion-extension creates a difference in body mass distribution, which aids in self-stabilization and improves gait control by changing leg angle and angular velocity previous to touchdown \citep{schilling}. As a result, additional speed caused by the spine increases the average speed of the animal \citep{hildebrand1960}. 

Since leg muscles are essentially velocity-limited actuators, additional energy provided apart from legs is important. At low-speed gaits, such as trotting, the spine reduces the energy carried by leg muscles and improves motion control \citep{schilling2010}. At high-speed gaits, according to Alexander, the intense energy needs of moving the legs at high speeds cannot be met solely by leg muscles. The additional required energy must come from the spine muscles \citep{alexander1981}. The spine muscles produce significant mechanical strength during spinal contraction, supplementing the power of the legs \citep{gray2013}. Additionally, bending of the spine during rear leg movement helps to propel the front legs more effectively than using just the muscles of the back legs alone \citep{hildebrand1959}.

Bertram et al discovered that spine motion could increase energy efficiency \citep{bertram}. In galloping quadrupeds, the spine has been stated as effective energy storage \citep{alexander1985}. Leg muscles and tendons have been considered like spring-damper structures that retain elastic energy and lower the energetic cost of dynamic movements \citep{alexander1984,alexander1990}. In addition to leg muscles and tendons, elastic structures, especially aponeurosis, which is a tendon-like system in the back that works like a spring, will regenerate some kinetic energy, minimizing the amount of energy required during high-speed running gaits \citep{alexander1984gaits}. In summary,  the musculoskeletal framework of a quadruped's spine functions like an elastic component, it retains elastic energy in the torso to limit energy intake \citep{roberts}.

In nature, the benefits of an integrated spine seem to be closely linked to gait selection. At slower speeds, quadrupedal animals use symmetrical gaits, in which the legs on the right side execute the very same action as the legs on the left, but 180 degrees out of phase. They prefer to switch to asymmetrical gaits at higher speeds, under which opposing sides of the animal execute distinct movements or the phase change varies from $180^{\circ}$ \citep{alexander1988,hoyt}. For these high-speed asymmetrical gaits, animals mainly use the articulation in their spine \citep{alexander1985,fischer}, while symmetrical gaits show less spine motion \citep{alexander1988}. Furthermore, these findings provide insight into the spine's role in gait transformation. This is a significant point, and any research into the utility of an articulated spine in legged robots should take gait into account.

In conclusion, creatures such as cheetahs and hounds use their spines to perform incredibly fast gaits. The spine movement raises the successful stride speed, gives auxiliary strength to the legs, and aids in the harnessing of energy by storing and releasing it, according to studies on these species. From an evolutionary standpoint, it's just normal to think of the spine as the animal body's propeller. Both of these findings point to the significance of the spine in locomotion.
Under these circumstances, the possibilities to boost the motion performance of the quadrupedal robots in the matter of speed, mobility, and energy is expanded. Thus, adding a spine to quadruped robots is an effective way to increase their performance. Since the spine aids in the reduction of ground reaction forces, while the compliant effect aids in the improvement of stability.

\subsection{Related Work}

Several quadrupedal robots have been built to take advantage of having an articulated body, which is a trait shared by animals. Cordes et al. created a four-legged walking robot with a flexible spine that allowed for more stable walking \citep{cordes}. A planar horse model was introduced by Herr and McMahon. Linear springs were also used to mimic back and neck movements in stance \citep{herr2000,herr2001}. Lewis and Bekey developed The GEO-II, a quadruped robot with a flexible spine \citep{lewis}. Literature on the quadrupedal robots with articulated spines can be reviewed under four subgroups such as comparative studies on the effects of various spine builds, effects of articulated spines on energy efficiency, gait and posture stabilization, and dynamic gait performance.

\subsubsection{Comparative studies}

A variety of studies have used comparative studies among quadrupedal models to determine the possible benefits of an articulated spine. Comparison between two models of a rigid body with and without two spinal spine joints revealed that the spinal joints would minimize the impact of ground reaction forces on the body in a specific velocity range \citep{li2020}.

Haueisen discussed two 2D quadrupedal bounding models, one with an articulated spine with 6 rigid parts and one with a rigid spine with 5 rigid parts. Haueisen used these models to investigate the influence of speed and stride frequency on the bounding gait's energy requirements. She discovered that the articulated model used the articulated spinal joint in similar ways to that seen in nature when bounding, as well as seeing kinetic advantages at faster speeds. As compared to a rigid body model, the cost of transport (the effort required to shift a unit weight over a unit distance) and the height change of the center of mass were lower for a model with spine joint \citep{haueisen}.

Pouya et al. used a planar simplified model to derive the dynamics of a quadruped robot with flexible spine motion. Open-loop bounding motions are evaluated by optimizing actuation profiles. They investigated the influence of flexible spine motion on stability \citep{pouya2012}. In the following study, they presented how spinal joint actuation and compliance affected the bounding efficiency of a virtual compliant Bobcat robot. They compared two quadruped versions, one with a passive spine and the other with an active spine, and discovered that only when the spine is actuated does decrease spinal stiffness lead to better energy production \citep{pouya2017}.

Eckert et al. proposed Lynx-robot, a miniature quadruped robot in order to compare the effects of the various spine and leg designs \citep{eckert2015}. They also designed Serval, a low-cost quadruped robot with an actuated spine, using engineering tools and interdisciplinary expertise translated from biology. In their research with Serval, they employed a variety of quadruped gaits \citep{eckert2020}.

\subsubsection{Energy efficiency}

According to some studies, the robot model with a spinal joint has the benefit of lowering transportation costs. Chen et al. discovered that spine motion, spinal stability, and energy conservation are all related concepts \citep{chen}. The addition of a soft structure to the spine creates a passive spine. The structural properties of the spine determine how it functions. In order to improve performance, the stiffness of the passive spine can be changed \citep{takuma}. Seipel explored the possibility of producing bounding motions without actuation using an entirely passive model. The model used in this study refers to the geometry of two spring-loaded inverted pendulums bound by a rotational spring, with the research taking place in reduced gravity \citep{seipel2011}. Deng et al. proposed a quasi-passive model in which bounding was accomplished by assuming that the torso joint can be ‘locked' at its maximal flexion and extension. Results indicated that adding a spine joint to the quasi-bounding model reduced the peak leg force while also lowering ground reaction forces \citep{deng}.

Leeser investigated the impact of an articulated spine in a planar quadruped robot in one of his early works. In order to examine the function of the spine in running, Leeser designed a planar quadruped robot with two driving joints in the spine. To absorb impacting energy, he added a spring to the end of each hydraulic leg actuator. Raibert's finite state machine bounding controller \citep{raibert1986} is employed in order to recreate motion sequences in which quadruped animals, such as cats, use the spine degree of freedom. The leg length of the robot is maintained until the air spring achieved its optimum compression at each stance. He discovered that the two spinal driving joints could help legs operate more efficiently, and that spine motion could change the impedance characteristic between the front and back legs in addition to that the spine-equipped robot could extend the effective length of a leg,  provide auxiliary power to legs, and aid in energy coordination \citep{leeser}.

Fanari, a 3D passive dynamic quadruped robot developed by Kani et al., can gallop down a slope without the use of external energy. They concluded that the passive elements of the spine's structure arrangement can be altered to enhance performance \citep{kani2011}. For a robotic simulation comparing rigid, passive articulated, and dynamic articulated spinal configurations, they also used a central pattern generator (CPG) controller to achieve bounding gaits. In terms of bounding power consumption, they discovered that a series elastic actuated spine outperformed an articulated passive spine and a rigid spine \citep{kani2013}.

Using reduced-order, energy conservative models, Cao and Poulakakis studied the impacts of a passive spine degree of freedom on the robot's bounding performance \citep{cao2012}. In the following study, they developed a dimensionless passively quadruped model, figuring out that the self-stability of quadruped bounding is aided by some stiffness combinations of leg and spine, greater moment of inertia around the mass-center of the spinal section, and a longer spinal segment. To produce the bounding gait, they used a passive and conservative model with a segmented flexible torso and compliant legs. For certain combinations of system parameters, it was discovered that a number of running velocities can be realized within the same overall energy level, and self-stable bounding motions can be found \citep{cao2013}. In addition to that they, using optimal control, discovered that an articulated model of a bounding quadruped was more economical than a rigid model at sufficiently high speeds by maximizing the energy distribution \citep{cao2014}. They also discovered that improved spinal stiffness contributes to a higher stride frequency and that a higher ratio of spinal mass to the total mass of a quadruped robot improves horizontal speed and reduces energy consumption \citep{cao2015}. 

Yesilevskiy et al. investigated the effects of an articulated spine through a variety of quadrupedal gaits. They used related models to show that a flexible-articulated spine can help quadruped robots achieve optimum speed and energy efficiency in asymmetrical gaits. Their findings revealed that an articulated spine increases maximum potential speed and enhances locomotor economy at higher speeds \citep{yesilevskiy}.

In another study effects of an asymmetric segmented body are investigated. The results show that an asymmetric segmented body has a larger spine oscillation, a shorter stride time, and a lower transport cost, all of which aid the robot's efficiency. In addition to this, the asymmetric segmented spine makes the quadruped robot function more efficiently than the symmetric segmented spine, according to Phan et al \citep{phan}.

\subsubsection{Gait and posture stabilization}

Various studies concentrated on the effects of an articulated spine on gait stabilization. For instance, According to Li et al., the commanded spine has a compliant impact on the contact with the ground in trotting gait, which is useful for enhancing stability \citep{li2020compliant}. Park et al. demonstrated a quadruped walking robot with a spine joint that has a discontinuous spinning gait. They used simulation to define a kinematic relationship between the spine joint, the hip, and the body's center of gravity, and then to determine the optimum values of parameters to spin the most stable \citep{park}. Aoi studied the impact of a compliant spine on a quadruped robot's gait transition. The locomotion of a quadruped robot with a roll joint between the front and rear bodies was studied in a simulation environment. They discovered a roll spine would cause the gait to transition from crawl to trot by changes in roll joint stiffness \citep{aoi}. A quadruped robot with a pneumatically actuated skeleton was presented by Tsujita and Miki. They conducted a feasibility analysis on gait pattern stability using pneumatic actuators with adjustable elasticity. The robot was able to perform stable locomotion in a variety of motion patterns \citep{tsujita}.

LightDog, a quadruped robot with a global compatible spine, was created by Zhang et al. They demonstrated that a flexible spine would increase balance and postural stability \citep{zhang2013}. In a later work of the team, the effect of non-flexible spine motion on stability was investigated too \citep{zhang2014}.

The quadruped robot's workspace volume is increased by Duperret et al.'s geared core actuation. They illustrated their research with a spined quadruped leaping on both smooth and uneven terrain \citep{duperret2016}. They also used an actively driven spine on the Inu quadrupedal robot to demonstrate empirically stable bounding, and they suggest a reduced-order model in order to capture the dynamics associated with this extra, actuated spine degree of freedom \citep{duperret2017}.
Nie et al. discovered that decreasing the front body-mass to rear body-mass ratio would improve quadruped bounding with an articulated spine locomotion efficiency without sacrificing motion stability \citep{nie}.

\subsubsection{Dynamic gait performance}

Various studies in literature concentrate on the effects of articulated spine on dynamic gait performances. Masuri et al. proposed a strategy for a quadrupedal robot with an active back joint to practice a self-learning dynamic walk. They increased walking efficiency by optimizing 12 of the robot's complex walking parameters, including active rear bending \citep{masuri}. Weinmeister et al. created the Cheetah-cub-S, a miniature quadruped robot with an active and flexible spine. The inclusion of the spine improves the robot's flexibility and maneuverability, according to the findings \citep{weinmeister}. Fisher et al. examined the impact of the spine on sudden acceleration. They employed quadruped robot sagittal simplified models to compare different spine morphologies in terms of stride average acceleration, and they discovered that the articulated spine is not superior to the prismatic spine in this regard \citep{fisher}. Furthermore, the number of spinal joints in quadruped motion has a significant impact on motion performance \citep{zhao2012}. In a simulation scenario, Bhattacharya et al. studied the effects of spinal joint compliance and actuation on bounding efficiency of a 16 degrees of freedom (DOF) quadruped robot Stoch 2 \citep{bhattacharya}.

For the quadruped robot with non-flexible spine motion, \c{C}ulha and Saranli suggested a planar simplified model. An actuated spine joint is present in the planar body. They investigated the impact of non-flexible spine motion on speed, especially how the simplified model could accomplish the bounding gait using the planar simplified model. They introduced a bounding gait controller and demonstrated how attaching a spine joint to a virtual quadruped robot would improve forward speed and hopping height. They used a bounding robot model based on a simplified leg model and a single DOF spinal joint. Bounding was generated using PID control loops that imposed desired values on the relative angle between the two torso segments. The results of the analysis between the rigid bounding model and the model with a spinal joint revealed that the actuated spine mechanism would lengthen the stride and enable the model to reach a higher hopping height \citep{culha2011}. In a later study for the quadruped robot with flexible spine motion, \c{C}ulha suggested a planar simplified model. There was an actuated spine joint in the planar simplified model's skeleton, as well as a linear torsion spring in the spine joint. The effect of flexible spine motion on speed was investigated using a simplified planar model \citep{culha2012}.

Zhao et al. investigated how spine articulation affects quadruped motion. They presented a quadruped robot that was strictly controlled by its spine and studied how non-flexible spine motion affects stability \citep{zhao2012embodiment}. After that study, they created a pneumatically powered quadruped "Kitty" with a rigid, passive, and active spine structure. They discovered that while the spinal motions are coordinated with leg movements, the active spine's extension and flexion enables the robot to attain higher speeds when using a step-function control pattern to achieve a bounding gait \citep{zhao2013}.

Wei et al. use a 12-DOF galloping model to investigate how the spinal joint affects galloping gait performance. The spinal joint model proposed in this study is operated by a lock-unlock process, in which the spinal joint is locked when a maximal extension or flexion is achieved. The effects of different initial angles of the spinal joint, different postures, and different horizontal velocities on the functional performance of dynamic models with and without a spine were investigated. Results present that gait performance is improved by reducing the ground reaction forces \citep{wei2015rotary}. In a later work, they presented two quadruped models, one with a rigid spine and the other with a passively flexible articulated spine, and found that the flexible spine reduces the center of mass vertical fluctuation, ground reaction forces, and energy consumption more than the rigid spine \citep{wei2015effect}.

Khoramshahi et al. built a quadruped robot with a spinal driving joint. They demonstrated that spine motion could improve system stability and energy efficiency through experiments with the robot. Multiple unconstrained flight-phase bounding gaits were demonstrated. The successful spine movement decreased the robot's horizontal instantaneous impulse during bounding \citep{khoramshahi}.

Folkertsma et al. constructed MIT Cheetah robot. They investigated a quadrupedal model with one spinal joint and four actuated legs, including hip and knee joints, in a rotary galloping gait. They created a mechanical arrangement for the quadruped robot's spine that allows for flexible spine motion. A differential gear controls the spine structure, which can adjust its stiffness. The spine joint of the MIT Cheetah robot was controlled by the differential gear drive output of its rear legs. If both rear legs are out of a trot-gait phase transition, it becomes active \citep{folkertsma}.

Pusey et al. suggested the Canid, a quadruped robot with an actuated spine and a linkage-based actuated leg for high-frequency gaits \citep{pusey}. Canid is also capable of performing a leap-from-rest operation \citep{duperret}.

Wang et al. studied the passive spine's compliant effect and used a locking and unlocking mechanism to increase the dynamic motion of the spine. A bio-inspired controller built on a CPG was proposed to understand the bounding gait of their SQBot, which is a quadruped robot with a spine joint \citep{wang2017}. In addition to that at a later study, the bounding gait of a mini Cheetah Robot with a passive spine joint was studied by them, too \citep{wang2020}.

Zeng et al. investigated the influence of center of mass location of a spinal segment on dynamic motion of a quadruped robot with a symmetric spine and a flexible joint \citep{zeng}.


\subsection{Our Contribution}
Dynamic simulation has developed into a strong appliance for analyzing complex behaviors and working on new control strategies, in the robotic research field. Creating a virtual world for robotic research can contribute testing and prototyping complicated robot tasks that are difficult to implement in the real world. With the help of simulation environments, researchers have the opportunity to fulfill the anterior section of the robotic studies before experimenting in the real world. Hence, constructing a solid simulation environment is vital in the robotic research field.

Prior quadrupedal structures with spine joints mostly utilized simplified two-legged planar robot in simulation studies \citep{culha2011,cao2015}. Simplified models are well-known methods for grasping the fundamentals of leg dynamics. However, critical gait characteristics like swing leg dynamics, actuator dynamics, and energy dissipation are missing. A broad simplification like considering legs are massless, widens the difference between simulation and practice, making the task of implementing simulation results in robotic hardware more challenging. Apart from these, inaccuracies in simulation and fundamentally flawed ground reactions both play a role in transferring imprecise data to actual hardware. Therefore, high fidelity physics engine based simulation software programs are started to be employed by researchers to simulate the robot and environment model to solve the above issues and create a more realistic model of the robot and environment \citep{todorov2012,koenig,michel,rohmer}.

In this paper, we provide a construction method of a spined quadruped robot simulation environment for conducting comprehensive studies on the effects of spinal joint actuation. A complex, physical robot model serves as the foundation for our simulation world. We developed a set of mathematical dynamic models for a quadruped robot with two DOF rotational joints on the torso, i.e., roll and pitch angles. The Lagrangian approach was used to derive the robot's equations of motion. There are stance and swing phases during the locomotion from the nature of the quadrupedal gait. A quadruped robot has dynamics with contact forces during stance and without contact forces during the swing phase. Looking from the complete locomotion perspective, robot dynamics include contact dynamics, non-contact dynamics, and transition between them. This concept is called hybrid dynamics. In this work, the system's hybrid dynamics were modeled with hard ground interaction (i.e., no penetration). There is a necessity for a solid simulation environment for a comprehensive analysis of the benefits of actuation at the spinal joint and accurate simulation environment will be extremely useful in guiding the design process and understanding the basic processes underlying the observed behaviors.

The paper is structured as follows: Section \ref{sec:2} is modeling kinematics and dynamics of the 20 DOF spined quadruped robot. In Section \ref{sec:3}, the simulation environment with an accurate contact model is described. The paper is concluded and future work is discussed in Section \ref{sec:4}.

\section{Modeling of a spined quadruped robot}
\label{sec:2}
In this section, there are descriptions of spined quadruped robot that is utilized in the simulation environment. The configuration of the quadruped robot is explained. Furthermore, kinematics and dynamics modeling of this robot is derived. 

\subsection{Kinematic arrangement of the quadruped robot}

\begin{figure}[!ht]
\begin{center}
\includegraphics[width=5cm, trim=0.2cm 0.2cm 0.3cm 0.3cm, clip=true,  angle=0]{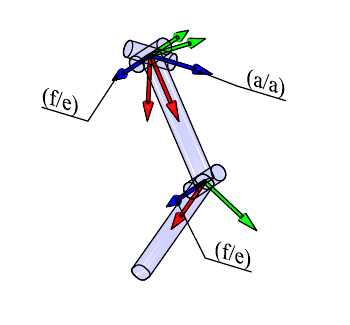}
\caption{The kinematic arrangement and the frame locations of the robot leg. $x$, $y$ and $z$ axes are represented with red, green and blue arrows, respectively.}
\label{fig1}
\end{center}
\end{figure}

\begin{figure}[!ht]
\begin{center}
\includegraphics[width=7cm, trim=0.2cm 0.2cm 0.3cm 0.3cm, clip=true,  angle=0]{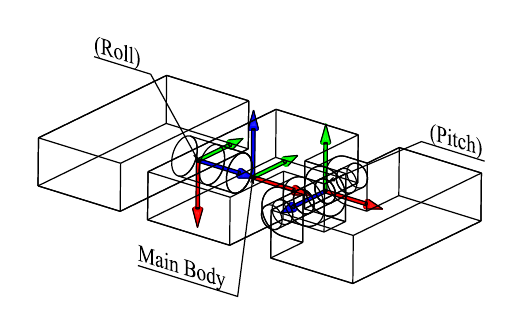}
\caption{The kinematic arrangement and the frame locations of the robot body. $x$, $y$ and $z$ axes are represented with red, green and blue arrows, respectively.}
\label{fig2}
\end{center}
\end{figure}

\begin{figure*}[!ht]
\begin{center}
\includegraphics[width=10cm, trim=0.2cm 0.2cm 0.3cm 0.3cm, clip=true,  angle=0]{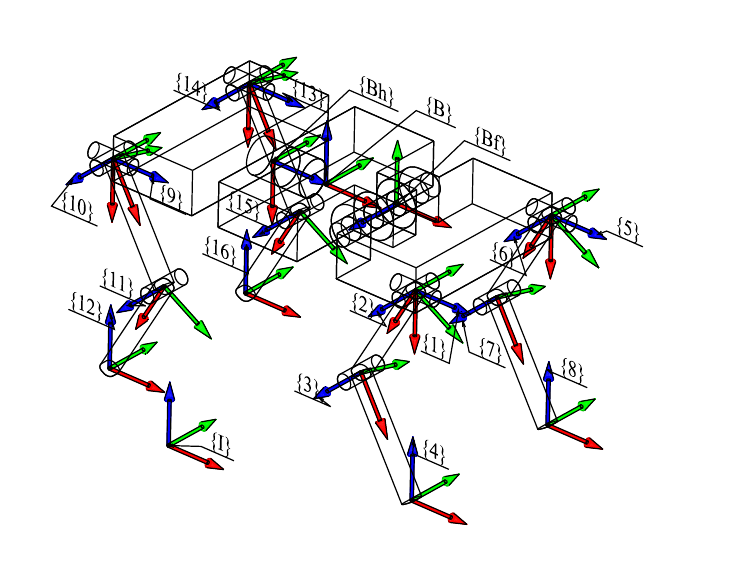}
\caption{The frame locations of the full-body quadruped. $x$, $y$ and $z$ axes are represented with red, green and blue arrows, respectively.}
\label{fig3}
\end{center}
\end{figure*}

Motion equations of a quadruped robot are derived by assigning a base-frame-located body center of mass (COM). The coordinate relationship among the inertial frame and the robot is described by a transformation matrix. The presented quadruped model consists of 20 DOF with 3 DOF on each leg and 2 DOF at the spine. The remaining 6 DOF represent floating-base in $3D$ space. Every DOF on legs and the body are rotational. Each leg has an adduction/abduction (a/a) joint on the hip, flexion/extension (f/e) joints on the hip and the knee. There is an illustration at Figure \ref{fig1}. All legs have identical frame attachments. There are three separate body parts and two joints to imitate active spine motion that can be seen in Figure \ref{fig2}. The schematic diagram of the full-body frame locations of the  quadruped is shown in Figures \ref{fig3}. ${I}$ is the inertial world frame. ${B}$ is the base frame fixed at the COM of body. Coordinate frames are attached to every link at the joint locations

Transformation matrices between the leg frames are in the following structure.
\begin{equation}
\label{eq1}
\begin{split}
    ^{i} _{i+1} T &= \left[\begin{array}{cc}
    { }^{i} _{i+1}R_{(3\times 3)} & { }^{i} r_{i,i+1(3\times 1)}\\
    0_{(1\times 3)} & 1_{(1\times 1)}
    \end{array}\right] \\
    &= \left[\begin{array}{cccc}
    c \theta_{i+1} & -s \theta_{i+1} & 0 & a_{i} \\
    s \theta_{i+1} c \alpha_{i} & c \theta_{i+1} c \alpha_{i} & -s \alpha_{i} & -s \alpha_{i} d_{i+1} \\
    s \theta_{i+1} s \alpha_{i} & c \theta_{i+1} s \alpha_{i} & c \alpha_{i} & c \alpha_{i} d_{i+1} \\
    0 & 0 & 0 & 1
    \end{array}\right],
\end{split}
\end{equation}
where $R$ is an orthogonal rotation matrix, $r$ is the position vector between frames and $a$, $\alpha$, $d$, $\theta$ are DH parameters \citep{denavit}. The DH parameters of each leg in the simulation model are shown in Table \ref{tab1}.

\begin{table}[!ht]
\caption{DH parameters of the legs in the simulation model.}
\begin{center}
\begin{tabular}{|c|c|c|c|c|c|}
\hline
Leg & Frame & $a_{i}$ & $\alpha_{i}$ & $d_{i+1}$ & $\theta_{i+1}$  \\
\hline
\multirow{4}{*}{Front Right} & 1 & 0 & 0 & 0 & $\theta_1 (a/a)$ \\
\cline{2-6}
& 2 & 0 & $-\pi/2$ & 0 & $\theta_2 (f/e)$ \\
\cline{2-6}
& 3 & $L_{fr_1}$ & 0 & 0 & $\theta_3 (f/e)$ \\
\cline{2-6}
& 4 & $L_{fr_2}$ & 0 & 0 & 0 \\
\hline
\multirow{4}{*}{Front Left} & 5 & 0 & 0 & 0 & $\theta_4(a/a)$ \\
\cline{2-6}
& 6 & 0 & $-\pi/2$ & 0 & $\theta_5(f/e)$ \\
\cline{2-6}
& 7 & $L_{fl_1}$ & 0 & 0 & $\theta_6(f/e)$ \\
\cline{2-6}
& 8 & $L_{fl_2}$ & 0 & 0 & 0 \\
\hline
\multirow{4}{*}{Hind Right} & 9 & 0 & 0 & 0 & $\theta_7(a/a)$ \\
\cline{2-6}
& 10 & 0 & $-\pi/2$ & 0 & $\theta_8(f/e)$ \\
\cline{2-6}
& 11 & $L_{hr_1}$ & 0 & 0 & $\theta_9(f/e)$ \\
\cline{2-6}
& 12 & $L_{hr_2}$ & 0 & 0 & 0 \\
\hline
\multirow{4}{*}{Hind Left} & 13 & 0 & 0 & 0 & $\theta_{10}(a/a)$ \\
\cline{2-6}
& 14 & 0 & $-\pi/2$ & 0 & $\theta_{11}(f/e)$ \\
\cline{2-6}
& 15 & $L_{hl_1}$ & 0 & 0 & $\theta_{12}(f/e)$ \\
\cline{2-6}
& 16 & $L_{hl_2}$ & 0 & 0 & 0 \\
\hline
\end{tabular}
\label{tab1}
\end{center}
\end{table}
The transformation matrix between the inertial frame and robot body frame is
\begin{equation}
\label{eq2}
\begin{split}
    ^{I} _{B} T &= \left[\begin{array}{cc}
    { }^{I} _{B}R_{(3\times 3)} & { }^{I} r_{I,B(3\times 1)}\\
    0_{(1\times 3)} & 1_{(1\times 1)}
    \end{array}\right],
\end{split}
\end{equation}
where ${ }^{I} _{B}R$ is an orthogonal rotation matrix representing the orientation of the robot body, ${ }^{I} r_{I,B}$ is the position vector between the inertial frame and the robot body COM, with respect to the inertial frame. The transformation matrices of the front and hind body frames that is located at the joint position of the links are as follows:
\begin{equation}
\label{eq3}
\begin{split}
    ^{B} _{B_f} T &=     \left[\begin{array}{cccc} \cos\left({\theta}_{B_f}\right) & -\sin\left({\theta}_{B_f}\right) & 0 & {L_b}/{2}\\ 0 & 0 & -1 & 0\\ \sin\left({\theta}_{B_f}\right) & \cos\left({\theta}_{B_f}\right) & 0 & 0\\ 0 & 0 & 0 & 1 \end{array}\right], \\
    ^{B} _{B_h} T &=     \left[\begin{array}{cccc} 0 & 0 & 1 & -{L_b}/{2}\\ \sin\left({\theta}_{B_h}\right) & \cos\left({\theta}_{B_h}\right) & 0 & 0\\ -\cos\left({\theta}_{B_h}\right) & \sin\left({\theta}_{B_h}\right) & 0 & 0\\ 0 & 0 & 0 & 1 \end{array}\right].
\end{split}
\end{equation}
Here $L_b$ is the length of the main body. $\theta_{B_f}$ is the pitch motion of the spine that is located between main body and front body. $\theta_{B_h}$ is the roll motion of the spine that is located between the main body and hind body.  Once and for all, the transformation matrices between the body links and the robot hips (leg connections) are given as:
\begin{equation}
\label{eq4}
\begin{split}
    ^{B_f} _{f_r} T &= \left[\begin{array}{cccc} 0 & 0 & 1 & {L}_{B_f}\\ -1 & 0 & 0 & -{H}_{B_f}/2\\ 0 & -1 & 0 & {W}_{B_f}/2\\ 0 & 0 & 0 & 1 \end{array}\right], \\
    ^{B_f} _{f_l} T &= \left[\begin{array}{cccc} 0 & 0 & 1 & {L}_{B_f}\\ -1 & 0 & 0 & -{H}_{B_f}/2\\ 0 & -1 & 0 & -{W}_{B_f}/2\\ 0 & 0 & 0 & 1 \end{array}\right], \\
    ^{B_h} _{h_r} T &= \left[\begin{array}{cccc} 1 & 0 & 0 & {H}_{B_h}/2\\ 0 & 1 & 0 & -{W}_{B_h}/2\\ 0 & 0 & 1 & -{L}_{B_h}\\ 0 & 0 & 0 & 1 \end{array}\right], \\
    ^{B_h} _{h_l} T &= \left[\begin{array}{cccc} 1 & 0 & 0 & {H}_{B_h}/2\\ 0 & 1 & 0 & {W}_{B_h}/2\\ 0 & 0 & 1 & -{L}_{B_h}\\ 0 & 0 & 0 & 1 \end{array}\right].
\end{split}
\end{equation}
where $L$, $H$ and $W$ are the front and rear body's length, height and width, respectively.

\subsection{Orientation of the robot body}
\label{sec:2.2}

Representing the angular position of the robot body in the $3D$ space is carried out as follows. Considering the orientation of the robot body has three degrees of freedom in $3D$ space, three principal rotations are unified for its description. The orientation of a rigid body can be described as the orientation of a fixed reference frame at body \citep{goldstein}. We can find a mapping matrix between the inertial frame and the frame attached at the main body of the quadruped robot. In this work, three successive rotations around the inertial frame are selected as roll, pitch and yaw rotations with the aim of describing the robot body orientation.

\begin{equation}
\label{eq5}
\begin{split}
    {}^{I}_{s_1}R_x(\phi) =& \begin{bmatrix} 
&    1 & 0 & 0 &\\ 
& 0 & \hspace{0.1cm} cos(\phi) & \hspace{0.1cm} -sin(\phi) &\\
& 0 & \hspace{0.1cm} sin(\phi) & \hspace{0.1cm}  cos(\phi) &
    \end{bmatrix}, \\ 
    {}^{s_1}_{s_2}R_y(\Theta)=& \begin{bmatrix} 
& cos(\Theta) & \hspace{0.1cm}  0 & \hspace{0.1cm}  sin(\Theta) &\\
&    0 & \hspace{0.1cm} 1 & 0 &\\ 
& -sin(\Theta) & \hspace{0.1cm}  0 & \hspace{0.1cm}  cos(\Theta) &
    \end{bmatrix}, \\
    {}^{s_2}_{B}R_z(\psi)=& \begin{bmatrix} 
& cos(\psi) & \hspace{0.1cm}  -sin(\psi) & \hspace{0.1cm}  0 &\\
& sin(\psi) & \hspace{0.1cm}  cos(\psi) & \hspace{0.1cm}  0 &\\
&    0 & \hspace{0.1cm}  0 & \hspace{0.1cm}  1 &
    \end{bmatrix}, \\     {}^{I}_{B}R(\phi,\Theta,\psi) = & {}^{I}_{s_1}R_x(\phi) \hspace{0.1cm} {}^{s_1}_{s_2}R_y(\Theta)\hspace{0.1cm} {}^{s_2}_{B}R_z(\psi).
    \end{split}
\end{equation}
Here $\phi$, $\Theta$ and $\psi$ are roll, pitch and yaw angles, respectively. ${}^{I}_{B}R$ is the rotation matrix between inertial frame and the robot body frame. $s_1$ and $s_2$ are sub-rotation frames. This formulation is traditionally named as the Euler Angles representation. Finding a mapping matrix between the rates of Euler angles and the angular velocity of the robot body is essential in the construction of a $3D$ robotic simulation environment. Using sub-frames the relation between the angular velocity of the robot body with respect to the body frame and Euler angle rates are derived as below.
\begin{equation}
\label{eq6}
    {}^{B}\omega_{B}= {}^{B}_{s_2}R \hspace{0.1cm} {}^{s_2}_{s_1}R \begin{bmatrix} 
    \dot{\phi} \\  0 \\  0 
    \end{bmatrix} + {}^{B}_{s_2}R \begin{bmatrix} 
    0 \\  \dot{\Theta} \\  0 
    \end{bmatrix} + \mathtt{I}_{(3\times{3})} \begin{bmatrix} 
    0 \\  0 \\  \dot{\psi} 
    \end{bmatrix}.
\end{equation}
After matrix operations, the mapping matrix is given as follows.
\begin{equation}
\label{eq7}
\begin{split}
    {}^{B}\omega_{B}= & E \Omega, \\
    E(\Theta,\psi) = & \begin{bmatrix} 
    cos(\Theta)cos(\psi) \hspace{0.1cm} & sin(\psi) \hspace{0.1cm} & 0 \\ -cos(\Theta)sin(\psi) \hspace{0.1cm} & cos(\psi) \hspace{0.1cm} & 0 \\ sin(\Theta) \hspace{0.1cm} & 0 \hspace{0.1cm} & 1
    \end{bmatrix}, \\ \Omega = & \begin{bmatrix} 
    \dot{\phi} \\  \dot{\Theta} \\ \dot{\psi} 
    \end{bmatrix}.
\end{split}
\end{equation}
Above, $E$ is the mapping matrix among body angular velocity with respect to body frame and Euler angle rates. Since Euler angles are not orthogonal, the inverse of the mapping matrix ($E$) is not equal to its transpose.

\begin{equation}
\label{eq8}
\begin{split}
     \Omega = & E^{-1} \hspace{0.1cm} {}^{B}_{I}R \hspace{0.1cm} {}^{I}\omega_{B}, \\ E_t(\phi,\Theta) = & \begin{bmatrix} 
    1 \hspace{0.1cm} & \frac{sin(\Theta)sin(\phi)}{cos(\Theta)} \hspace{0.1cm} & \frac{-sin(\Theta)cos(\phi)}{cos(\Theta)} \\ 0 \hspace{0.1cm} & cos(\phi) \hspace{0.1cm} & sin(\phi) \\ 0 \hspace{0.1cm} & \frac{-sin(\phi)}{cos(\Theta)} \hspace{0.1cm} & \frac{cos(\phi)}{cos(\Theta)}
    \end{bmatrix}
\end{split}     
\end{equation}
where $E_t$ is a mapping matrix (i.e., $E_t=E^{-1} \hspace{0.01cm} {}^{B}_{I}R$) between Euler angle rates and angular velocity of the robot body with respect to inertial frame. It can be observed that the mapping matrix becomes singular when pitch angle ($\Theta$) equals $90^\circ$ or $270^\circ$. However, it is not likely to have these values at the quadruped locomotion. Nevertheless, a very small overflow number could be added in order to overcome singularity.

Formally, a robot jacobian relation is a set of partial differential equations related to robot joints. Furthermore, a robot jacobian defines relation between velocities of the configuration space and the task space.

\begin{equation}
\label{eq9}
\begin{split}
    \frac{\partial x}{\partial t} = \frac{\partial x}{\partial q} \frac{\partial q}{\partial t} \hspace{0.2cm} \Rightarrow \hspace{0.2cm} \dot{x} = J \dot{q}
\end{split}     
\end{equation}
However, floating-base robots include the body orientation in their generalized coordinates ($q$). If orientation of the robot is represented with Euler angles (as in this work), the $\frac{\partial q}{\partial t}$ term contains Euler angles rates instead of angular velocity of the robot body. Hence, a correction must be applied to $\frac{\partial x}{\partial q}$ with the mapping matrix ($E_t$) in order to find an accurate jacobian matrix. Since there are several differential operations in the Lagrangian method, the corrections with the mapping matrix ($E_t$) are utilized in generating dynamic equations as well.

\subsection{Floating-base robot dynamics}
\label{sec:2.3}

A quadruped robot is modelled as a floating-base robot that has no fixed point in space but interacts with the environment. The quadruped robot dynamics are considered as a combination of generalized robot coordinates with a floating-base. The generalized robot vectors are

\begin{equation}
\label{eq10}
    q = 
    \begin{bmatrix} 
    q_{b}^T \hspace{0.25cm} q_{j}^T 
    \end{bmatrix}^T,
    \dot{q} = 
    \begin{bmatrix} 
    v_{b}^T \hspace{0.25cm} \dot{q}_{j}^T 
    \end{bmatrix}^T,
    \ddot{q} = 
    \begin{bmatrix} 
    a_{b}^T \hspace{0.25cm} \ddot{q}_{j}^T 
    \end{bmatrix}^T,
\end{equation}
where $q_{b}\in{SE(3)}$ corresponds to the position and orientation of the robot body with respect to an inertial frame. $q_{j}\in\mathbb{R}^{14}$ is the vector of joint angular positions of the quadruped robot with $14$ joints. $v_{b}\in\mathbb{R}^6$ and $a_{b}\in\mathbb{R}^6$ are spatial velocity and acceleration of the body, respectively. The motion equations of quadruped robot in contact with the environment are expressed as:

\begin{equation}
\label{eq11}
    M(q)\ddot{q}+C(q,\dot{q})+G(q) = S^T \tau + J_C(q)^T F_C,
\end{equation}
where $M(q)\in\mathbb{R}^{(6+14)\times{(6+14)}}$ is the inertia matrix, $C(q,\dot{q})\in\mathbb{R}^{(6+14)}$ is the Coriolis and centrifugal forces, $G(q)\in\mathbb{R}^{(6+14)}$ is the gravitational force, $J_C(q)\in\mathbb{R}^{(24)\times{(6+14)}}$ is the contact jacobian with respect to the inertial frame, $F_C\in\mathbb{R}^{24}$ is the vector of contact forces. $S\in\mathbb{R}^{(14)\times{(6+14)}}$ is a selection matrix of the actuated joints. Defining a selection matrix is beneficial during modeling floating-base robots, since the torso is not actuated. Hence, $S$ matrix is $\begin{bmatrix} 0_{14\times 6} \hspace{0.25cm} I_{14\times14} \end{bmatrix}$. $\tau\in\mathbb{R}^{14}$ is the vector of actuated joint torques.

The dynamic equations of a multi-body system can be developed in a variety of ways \citep{featherstone, kane}. All algorithms produce coequal set of equations, however, certain structures of equations can be more convenient for distinct objectives. In this work, the Lagrangian approach is chosen for acquiring dynamic equations. This method considers the energy properties of multi-body systems to derive the equations of motion. As a consequence, we will obtain all unknown matrices in (\ref{eq11}).

\subsection{Euler-Lagrange equations}
\label{sec:2.4}

This section is composed of the dynamic equations of the quadruped robot. The kinematic equations define the robot motion at velocity level. The dynamic equations indicate the relation among force and motion. Dynamic equations are prominent in robotic research topics such as robot design, simulation environment construction, and the design of control algorithms.

There are several methods in the robotic field for acquiring dynamic equations. The Lagrangian analysis is one of these methods. All methods constitute equivalent sets of equations. However, distinct equation forms may be preferable according to the task definition of the user. The Lagrangian approach exploits the Euler-Lagrange equations, which are derived by the kinetic and potential energy of the system. For the purpose of generating dynamic equations, a function called the Lagrangian of the system is defined as the difference among the kinetic and potential energy.

\begin{equation}
\label{eq12}
    L(q,\dot{q})= K(q,\dot{q}) - P(q).
\end{equation}
Here $K$ and $P$ denote the kinetic and potential energy of the system in generalized coordinates, respectively. The motion equations of a quadruped robot with generalized coordinates ($q\in\mathbb{R}^{20}$) and Lagrangian ($L$) are given by

\begin{equation}
\label{eq13}
    \frac{d}{dt}\frac{\partial L}{\partial \dot{q}_i} - \frac{\partial L}{\partial q_i} = F_i \hspace{0.5cm} i=1,...,n,
\end{equation}
where $F$ represents all external forces and torques acting on the body and links of the quadruped robot. $n$ is the size of the generalized joint coordinates. It equals $20$ in our quadruped robot. The equations in (\ref{eq13}) are called Euler-Lagrange equations. With the aim of applying Euler-Lagrange equations to a quadruped robot, the kinetic and potential energy of the robot links are stated as a function of the generalized positions and velocities. Considering every individual robot link is a rigid body, their kinetic and potential energy are described in terms of mass and moments of inertia about the link COM. The total kinetic energy of each link is given by the summation of translational and rotational kinetic energy.

\begin{equation}
\label{eq14}
\begin{split}
&    K_i=\frac{1}{2}V_i^T{}^i\mathbb{M}_iV_i \hspace{0.3cm} and \hspace{0.3cm} {}^i\mathbb{M}_i=\begin{bmatrix} 
&    m_i\mathtt{I}_{(3\times{3})} \hspace{0.1cm} & 0_{(3\times{3})}  &\\
&    0_{(3\times{3})}  \hspace{0.1cm} & {}^iI_i &
    \end{bmatrix}, \\ & i=1,...,d.
\end{split}
\end{equation}
Here $V_i\in{SE(3)}$ and $d$ denote the link spatial velocity and the total number of the link, respectively.  ${}^i\mathbb{M}_i\in\mathbb{R}^{6\times{6}}$ is the generalized inertia matrix of the link, expressed in the link frame. $\mathbb{M}$ is a symmetric and positive definite matrix. All of the links are modeled as a rectangular solid with mass $m$, length $l$, width $w$, and height $h$. The inertia tensor of each link is evaluated using the matrix below.

\begin{equation}
\label{eq15}
\begin{split}
&   {}^iI_i=\begin{bmatrix} 
&    \frac{m_i}{12}(w_i^2+h_i^2) & 0 & 0 &\\
&    0  & \frac{m_i}{12}(l_i^2+h_i^2) & 0 &\\
&    0  & 0 & \frac{m_i}{12}(l_i^2+w_i^2) &
    \end{bmatrix}, \\ & i=1,...,d.
\end{split}
\end{equation}
Since the coordinate axes are aligned with the principal axes of the rectangular solid, the inertia tensor is diagonal. However, in order to produce motion equations for floating-base robots, the inertia tensor must be expressed in the inertial frame. The instantaneous inertia tensor in relation to the inertial frame is evaluated by the configuration of an object. Hence, the inertia tensor of each link with respect to the inertial frame is computed by

\begin{equation}
\label{eq16}
    I_i={}^IR_i {}^iI_i {}^I{R_i}^T, \hspace{0.5cm} i=1,...,d,
\end{equation}
where ${}^IR_i$ is the rotation matrix between the frames. The generalized inertia matrix of each link with respect to inertial frame is given by

\begin{equation}
\label{eq17}
\mathbb{M}_i=\begin{bmatrix} 
&    m\mathtt{I}_{(3\times{3})} \hspace{0.1cm} & 0_{(3\times{3})}  &\\
&    0_{(3\times{3})}  \hspace{0.1cm} & I_i &
    \end{bmatrix}, \hspace{0.5cm} i=1,...,d.
\end{equation}
The total kinetic energy of the link $i$ is expressed with joint velocities as

\begin{equation}
\label{eq18}
    K_i(q,\dot{q}) = \frac{1}{2}\dot{q}^T J_i^T(q) \mathbb{M}_i J_i(q) \dot{q},
\end{equation}
where $J_i(q)\in\mathbb{R}^{6\times{20}}$ is the jacobian matrix of the link $i$ with respect to the inertial frame. The total kinetic energy of the quadruped robot is a summation of all links' kinetic energy:

\begin{equation}
\label{eq19}
    K(q,\dot{q}) = \sum^{d}_{i=1} K_i(q,\dot{q}) = \frac{1}{2}\dot{q}^T M \dot{q}.
\end{equation}
Here, $M\in\mathbb{R}^{20\times{20}}$ is the robot inertia matrix. The robot inertia matrix is created in terms of link jacobian matrices (COM of the links), $J_i$, and link generalized inertia matrices $\mathbb{M}_i$, by combining (\ref{eq18}), and (\ref{eq19}):

\begin{equation}
\label{eq20}
    M(q)= \sum^{d}_{i=1} J_i^T(q) \mathbb{M}_i J_i(q).
\end{equation}
The potential energy of the ith link is

\begin{equation}
\label{eq21}
    P_i(q) = m_igh_i(q),
\end{equation}
where the mass of the $ith$ link is $m_i$, and the gravitational acceleration is $g$. $h_i(q)$ is the position of the link COM with respect to inertial frame in the opposite direction of the gravitational acceleration. The total potential energy of the quadruped robot is a summation of all links potential energy:

\begin{equation}
\label{eq22}
    P(q) = \sum^{d}_{i=1}P_i(q) = \sum^{d}_{i=1}m_igh_i(q).
\end{equation}
When the potential energy (\ref{eq22}) is paired with the kinetic energy (\ref{eq19}), the Lagrangian of the quadruped robot is defined in terms of the generalized positions and velocities.

\begin{equation}
\label{eq23}
    L(q,\dot{q}) = \sum^{d}_{i=1}\left(K_i(q,\dot{q}) - P_i(q)\right) = \frac{1}{2}\dot{q}^T M \dot{q} - P(q).
\end{equation}
It would be convenient to express the kinetic energy of the quadruped robot as a summation form of a matrix-vector product, before applying the Lagrangian of the system into Euler-Lagrange equations.

\begin{equation}
\label{eq24}
    L(q,\dot{q}) = \frac{1}{2} \sum^{n}_{i}M_{ij} \dot{q}_i \dot{q}_j - P(q).
\end{equation}
The motion equations of the quadruped robot are obtained by substituting (\ref{eq24}) into Euler-Lagrange equations (\ref{eq13}) as,

\begin{equation}
\label{eq25}
\begin{aligned}
    & \frac{d}{dt}\frac{\partial L}{\partial \dot{q}_i} =  \frac{d}{dt}\left(\sum^{n}_{j=1}M_{ij} \dot{q}_j\right) = \sum^{n}_{j=1}\left(M_{ij} \ddot{q}_j + \dot{M}_{ij}\dot{q}_i \right) \\
    & \frac{\partial L}{\partial q_i} = \frac{1}{2}  \sum^{n}_{j,k=1}\frac{\partial M_{kj}}{\partial q_i} \dot{q}_k  \dot{q}_j - \frac{\partial P}{\partial q_i}, \hspace{0.5cm} i=1,...,n.
\end{aligned}
\end{equation}
The $\dot{M}_{ij}$ is expanded in terms of partial derivatives to reach the final form of the motion equations:

\begin{equation}
\label{eq26}
\begin{split}
    & \sum^{n}_{j=1}M_{ij}\ddot{q}_j + \sum^{n}_{j,k=1} \left(\frac{\partial M_{ij}}{\partial q_k} - \frac{1}{2}\frac{\partial M_{kj}}{\partial q_i} \right) \dot{q}_j\dot{q}_k + \frac{\partial P}{\partial q_i} = F_i \\ & i=1,...,n.
\end{split}
\end{equation}
If this motion equation is organised as the form given in (\ref{eq11}), the inertia  matrix, external forces, the Coriolis and centrifugal forces, and the gravitational force are obtained as

\begin{equation}
\label{eq27}
\begin{split}
    & M(q)= \sum^{d}_{s=1} J_s^T(q) \mathbb{M}_s J_s(q), \\
    & F= S^T \tau + J_C(q)^T F_C, \\
    & C_i(q,\dot{q}) =  \sum^{n}_{j,k=1} \left(\frac{\partial M_{ij}}{\partial q_k} - \frac{1}{2}\frac{\partial M_{kj}}{\partial q_i} \right) \dot{q}_j\dot{q}_k, \\ & G_i(q) =\frac{\partial P}{\partial q_i}, \hspace{0.5cm} i=1,...,n. 
\end{split}
\end{equation}
For the sake of getting accurate Coriolis, centrifugal and gravitational forces, a correction must be employed with the mapping matrix ($E_t$) as mentioned in Section \ref{sec:2.2}. 

\section{Simulation environment}
\label{sec:3}
Dynamic robot simulation is an important instrument that accompanies the robotic research field. It has the ability of testing complex actions before applying real robot hardware. The fundamental elements of the legged robotic simulation are the dynamic model of the robot and the contact environment. An elementary legged-robot simulator must include these features. Furthermore, the simulation environment should allow users to enrich their simulations with user-defined features such as reference generation, controller design, and optimization etc. Hence, users can develop their research by utilizing a simulation environment in their experiments.

\begin{figure*}[!ht]
\begin{center}
\includegraphics[width=14cm, trim=0cm 0cm 0cm 0cm, clip=true,  angle=0]{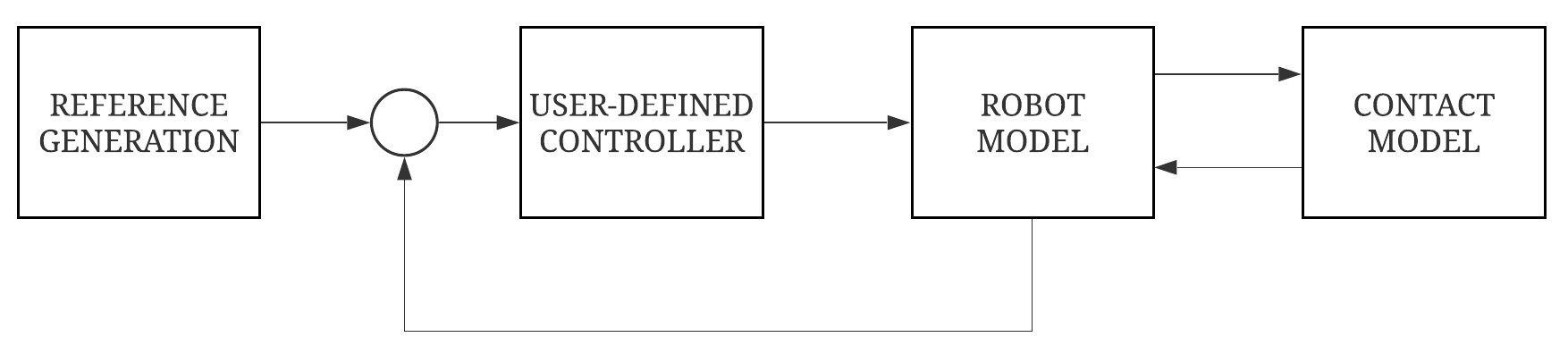}
\caption{The schematic representation of the simulation environment.}
\label{fig4}
\end{center}
\end{figure*}

The construction of the legged-robot simulation environment has two principal adversities. The first one is generating precise and accurate dynamics algorithms of the simulated robot and the environment and the other one is improving fast and efficient numerical solvers for computing their solutions. Considering these reasons, we developed our simulation environment in MATLAB/Simulink. MATLAB/Simulink supplies a suitable and resilient environment in integrating controllers, through its rich library and powerful solvers. Accordingly, MATLAB/Simulink interface provides suitable opportunities for robotic simulations. The generation of dynamic equations of the robot model is already mentioned in \ref{sec:2.4}. Increasing the computational speed of robot equations and generating an accurate dynamic contact model is mentioned in further. The schematic representation of the simulation environment is illustrated in Figure \ref{fig4}.

\subsection{Robot Model}
\label{3.1}
Robot motion equations are modelled in \ref{sec:2.4}. Nevertheless, they must be computed at reasonable times for the sake of a valid simulation environment. Due to analytical derivative operations in Lagrange analysis, symbolic calculations are convenient for generating matrices to be contained in dynamic equations of the robot. Symbolic functions and symbolic matrices are evaluated by $"subs()"$ and $"double(subs())"$ commands, respectively. However, these commands are not efficient in calculations of 20 DOF quadruped robot. There is a benefit generating $matlabFunctionBlock$ \citep{matlabfunction} in order to increase the speed of the evaluation time. When a symbolic function is converted to a $matlabFunctionBlock$, it optimizes the code utilizing intermediate variables named $t0$, $t1$, $t2$, etc. Furthermore, it can compose the matrices as a sparse numeric matrix by defining solely nonzero elements and picking up all other elements that are zeros. Since individual leg generalized coordinates of the quadruped robot are decoupled of other legs, there are large amounts of zero elements in the matrices. Hence, it is convenient to reshape them by employing sparse numeric matrix features. It not only simplifies matrices, but also decreases computation duration. Computation times of the matrices with $"double(subs())"$ and $matlabFunctionBlock$ are compared in Table \ref{tab2}. Computation times were measured with a computer that has $3~GHz$ processor and $128~GB$ RAM.

\begin{table}[!ht]
\caption{Computation duration of robot matrices.}
\begin{center}
\begin{tabular}{| m{2cm} | m{1cm}| m{1cm} |  m{1cm} | m{1cm} |}
\hline
Evaluation Method & $M(q)$ & $C(q,\dot{q})$ & $G(q)$ & $J_c(q)$   \\
\hline
Symbolic (subs command) & $8.061~s$ & $47.124~s$ & $0.246~s$ & $0.784~s$   \\
\hline
Optimized (matlab function) & $0.041~s$ & $0.107~s$ & $0.01~s$ & $0.01~s$   \\
\hline
\end{tabular}
\label{tab2}
\end{center}
\end{table}

\subsection{Contact model}
\label{sec:3.2}
Most of the former initiatives were based on spring-damper models. Although they are easy to implement, they lack stability and accuracy even with significant manual tuning \citep{adak}. Researchers began to utilize constraint-based contact physics instead of spring-damper model. These contact models were developed as a linear complementarity problem (LCP) \citep{yamane} and presented better results. Still, it is important to get a solution that is both physically accurate and computationally efficient. The aim is to boost both accuracy and efficiency. Being inspired by contact model solver \citep{todorov2010} of MuJoCo \citep{todorov2012}, we generate an accurate contact model to be utilized in our simulation environment.

Let us recall the motion equations of a quadruped robot from (\ref{eq11}). There is a possibility of reaching no solution in these motion equations due to Coulomb friction. Stewart and Trinkle presented a solution to this problem by representing the robot dynamics in discrete time \citep{stewart}. Let $h$ be the time step and $k$ be the time index. Replacing $\ddot{q}$ with $(\dot{q}_{k+h}-\dot{q}_{k})/h$, then (\ref{eq11}) and (\ref{eq9}) become:

\begin{equation}
\label{eq28}
\begin{split}
    M(q_k)\dot{q}_{k+h} = & M(q_k)\dot{q}_k + h (S^T \tau - C(q_k,\dot{q}_k) \\ & - G(q_k) + J_C(q_k)^T F_{C(k+h)}), \\ J_C(q_k) \dot{q}_{k+h} = & V_{k+h}.
\end{split}
\end{equation}
$\dot{q}_{k+h}$ can be eliminated by using the knowledge of $M$ is always invertible. After combining these two equations, we obtain the end-effector velocity as,

\begin{equation}
\label{eq29}
\begin{split}
    V_{k+h} = & J_C \left(\dot{q}_{k} + M^{-1}h(S^T \tau - C - G)\right) \\ & +J_CM^{-1}h J_C^T F_{C(k+h)},
\end{split}
\end{equation}
and this equation is organised in more compact form,

\begin{equation}
\label{eq30}
    A\lambda + V_0 = V.
\end{equation}
Here $A = J_CM^{-1} J_C^T $ is the inverse inertia matrix in the contact space, $\lambda = hF_{C(k+h)}$ is the contact impulse and $V_0 = J_C \left(\dot{q}_{k} + M^{-1}h(S^T \tau - C - G)\right)$ is the end-effector velocity when there is no contact. For transparency, $k$ index is suppressed in further derivations. $\lambda$ and $V$ is divided into normal and tangential components for utilizing in contact constraints. $\lambda_N\in\mathbb{R}$ and $V_N\in\mathbb{R}$ are denoted as normal components, while $\lambda_T\in\mathbb{R}^2$ and $V_T\in\mathbb{R}^2$ are denoted as tangential components. Constraints of normal components are defined as follows,
\begin{equation}
\label{eq31}
    \lambda_N \geq 0, \hspace{0.5cm} V_N \geq 0, \hspace{0.5cm} \lambda_N V_N = 0.
\end{equation}
The first constraint means that the contact impulse cannot pull to the ground. The second is that the robot foot cannot penetrate inside the ground, and the last one is that there is not any contact impulse unless the contact occurs. The constraints of tangent components are,
\begin{equation}
\begin{split}
\label{eq32}
    & \lambda_T \parallel V_T, \hspace{0.5cm} \lambda_T^TV_T\leq 0, \hspace{0.5cm} \lVert\lambda_T\rVert \leq \mu\lambda_N.
\end{split}
\end{equation}
These three constraints represent that the friction force and the foot velocity are parallel to each other, if slip emerges then the friction force must apply in opposite direction according to the slip velocity, and the friction force must stay inside of the friction cone. Here $\mu$ is the friction coefficient. 

Constraints in (\ref{eq31}) are called complementarity constraints. They compose a linear complementarity problem cooperatively with (\ref{eq30}). However, constraints in (\ref{eq32}) disrupt the LCP, as it contains nonlinear states i.e., slippage possibility. Therefore, the problem is converted into an unconstrained nonlinear optimization problem similar. The constraints defined in (\ref{eq31}) and (\ref{eq32}) are modified to nonlinear functions as follows,
\begin{equation}
\begin{split}
\label{eq33}
    \lambda_N(x) = & \max(0,-x_N), \\ V_N(x) = & \max(0,x_N), \\
    s(x) = & \min \left(1,\frac{\mu \lambda_N(x)}{\lVert x_T\rVert}\right), \\ \lambda_T(x) = & -s(x)x_T, \\ V_T(x) = & x_T - s(x)x_T. 
\end{split}
\end{equation}
Here $x_N$ and $x_T$ act like decision parameters of the contact model. Since only one of the $\lambda_N$ or $V_N$ can be non-zero (they are complementary), we can codify them with only one scalar ($x_N$). When $x_N$ is positive there is non-zero $V_N$, while there is non-zero $\lambda_N$ when $x_N$ is negative. Although the unit of the $x_N$ is changing with respect to its sign, it is not affecting mathematical computations. In the tangential plane, both $\lambda_T$ when $x_T$ can be non-zero (they are not complementary) at the same time. Still, they can be codified with one scalar ($x_T$) using their common direction, since these two vectors are always parallel to each other. Here the $s(x)$ function defines slippage condition. It gets the value of $1$ when there is no slippage. It is convenient to combine normal and tangential components of the contact impulse and foot velocity.

\begin{equation}
\begin{split}
\label{eq34}
    S(x)=& \begin{bmatrix} 
& 1 & \hspace{0.1cm} 0 & \hspace{0.1cm} 0 &\\
& 0 & \hspace{0.1cm}  s(x) & \hspace{0.1cm}  0 &\\
&    0 & \hspace{0.1cm}  0 & \hspace{0.1cm}  s(x) &
    \end{bmatrix}, \\ \lambda(x)= & -S(x)x, \hspace{0.5cm} V(x)= x-S(x)x.
\end{split}
\end{equation}
Here the contact impulse vector $\lambda(x)=\begin{bmatrix} \lambda_N(x) \hspace{0.25cm} \lambda_T(x) \end{bmatrix}^T$, foot velocity vector $V(x)=\begin{bmatrix} V_N(x) \hspace{0.25cm} V_T(x) \end{bmatrix}^T$, and x parameters vector equals $x=\begin{bmatrix} x_N \hspace{0.25cm} x_T \end{bmatrix}^T$. We can compose (\ref{eq33}) more organized as following,

\begin{equation}
\label{eq35}
    V(x) = \lambda(x) + x.
\end{equation}
Note that this equation provides all constraints specified in (\ref{eq33}). Since we are interested in finding only contact impulses, it is helpful defining relation of the foot velocity and the contact impulses by including all constraints. The contact impulse function can be defined with variable $x$ by inserting (\ref{eq30}) into (\ref{eq35}),

\begin{equation}
\label{eq36}
    \lambda(x) = (A-\mathtt{I})^{-1}(x - V_0),
\end{equation}
This nonlinear contact impulse equation can be solved by being adapted to unconstrained nonlinear optimization problem.

\begin{equation}
\label{eq37}
\begin{split}
& \underset{\textbf{x}}
{\text{min}} 
 \lVert(A-\mathtt{I})\lambda(x) - x + V_0\rVert^2,
\end{split}
\end{equation}
Although this optimization problem  finds the exact solution almost all the time, it cannot always meet to the exact solution. However, it still converges to a rational solution.

\section{Conclusion and future work}
\label{sec:4}

In this paper, we presented a broad review about quadruped robots with a spine joint. We classified spined quadruped robot research in the literature with main headings such as comparative studies on the effects of various spine builds, effects of articulated spines on energy efficiency, gait and posture stabilization, and dynamic gait performance to enlighten our future research directions. 

Kinematic and dynamic modeling of a 20 DOF floating-base robot is created to achieve complete motion equations of a quadruped with articulated 2 DOF spine. In order to achieve a more realistic robot model, legs are not modelled as massless since they contain approximately 25 percent of the overall mass. Euler-Lagrange equations are employed to reach the robot matrices in the general robot equation.

It is significant to get precise and accurate dynamics algorithms of the simulated robot and environment with fast and efficient numeric solvers in a simulation environment. Further, achieving precise simulation results in a long time is futile, since they cannot be employed on real-time applications. In order to prevent time inefficiency, optimized MATLAB/Simulink functions are applied to the general robot matrices. The solidity of the simulation environment is amplified with an accurate LCP based contact model rather than a spring-damper based contact model in order to increase model stability and accuracy. The employed contact model almost always reaches the exact solution, and it still converges to a rational solution when the exact solution cannot be found.

A presented simulation environment allows contributing further improvements of spined-based locomotion research mentioned in the related work. Our aim is to enrich this field in introduced aspects. Furthermore, we consider as a targeted future work to strengthen our simulation model with passive compliance computations.


%
%

\bibliographystyle{spbasic}      


%
%
\bibliography{template.bib}

\end{document}